\documentclass[11pt,a4paper]{article}
\usepackage[hyperref]{emnlp2018}
\usepackage{times}
\usepackage{latexsym}
\usepackage{url}
\usepackage{graphicx}
\usepackage{subfigure}
\usepackage{caption}
\usepackage{makecell}
\usepackage{ragged2e}
\usepackage{booktabs}
\usepackage{mathtools}

\graphicspath{ {images/} }

\aclfinalcopy 
\def\aclpaperid{392} 

\newcolumntype{L}[1]{>{\raggedright\let\newline\\\arraybackslash\hspace{0pt}}m{#1}}

\title{PreCo: A Large-scale Dataset in Preschool Vocabulary\\for Coreference Resolution}

\author{
Hong Chen\textsuperscript{1},
Zhenhua Fan\textsuperscript{1, 2},
Hao Lu\textsuperscript{1},
Alan L. Yuille\textsuperscript{3} and
Shu Rong\textsuperscript{1} \\
\textsuperscript{1}Preschool Lab, Yitu Tech \\
\textsuperscript{2}Shandong Normal University \\
\textsuperscript{3}Department of Cognitive Science and Department of Computer Science, \\
Johns Hopkins University \\
{\tt \{hong.chen,zhenhua.fan,hao.lv,shu.rong\}@yitu-inc.com} \\
{\tt ayuille1@jhu.edu}\\
}

\date{}

\begin{document}
\maketitle
\begin{abstract}
We introduce PreCo, a large-scale English dataset for coreference resolution. The dataset is designed to embody the core challenges in coreference, such as entity representation, by alleviating the challenge of low overlap between training and test sets and enabling separated analysis of mention detection and mention clustering. To strengthen the training-test overlap, we collect a large corpus of about 38K documents and 12.4M words which are mostly from the vocabulary of English-speaking preschoolers. Experiments show that with higher training-test overlap, error analysis on PreCo is more efficient than the one on OntoNotes, a popular existing dataset. Furthermore, we annotate singleton mentions making it possible for the first time to quantify the influence that a mention detector makes on coreference resolution performance. The dataset is freely available at \url{https://preschool-lab.github.io/PreCo/}.
\end{abstract}

\section{Introduction} \label{sec:introduction}

\begin{figure*}[htbp]
\centering
\includegraphics[width=15cm]{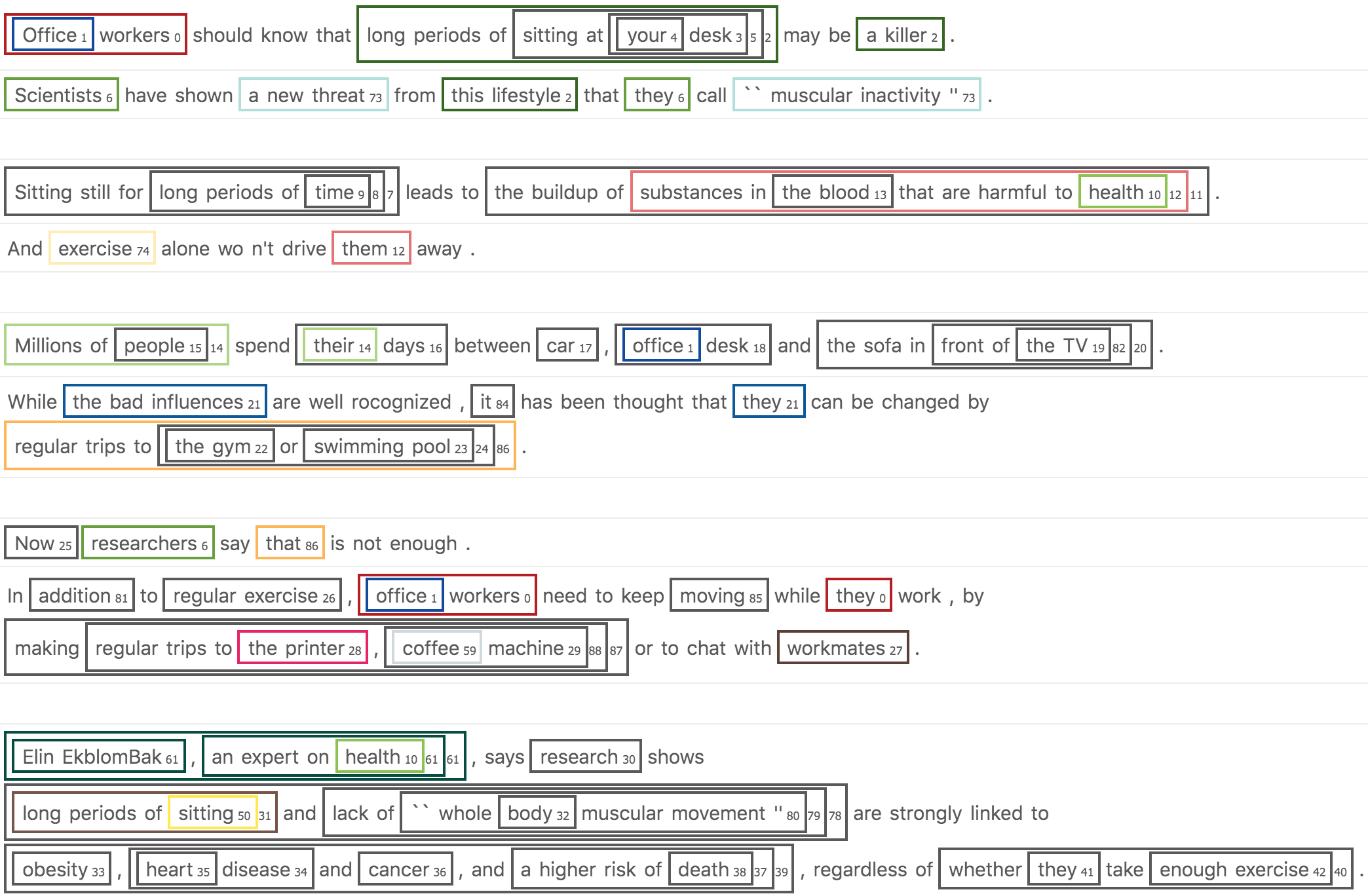}
\caption{An Example from PreCo. In the example, mentions are indicated by boxes, and mention clustering is indicated by the subscripted numbers. If two mentions have the same number, they refer to the same entity.}
\label{fig:example_text}
\end{figure*}

Coreference resolution, identifying mentions that refer to the same entities, is an important NLP problem. Resolving coreference is critical for many downstream applications, such as reading comprehension, translation, and text summarization.
Identifying a mention depends not only on its lexicons but also its contexts, and requires representations of all the entities before the mention. This is still a challenging task for the approaches based on the cutting-edge word2vec-like lexical representation. For example, it is hard to identify the mention ``he'' between two entities ``Tom'' and ``Jerry'' because they have almost the same word embeddings.

A number of datasets have been proposed to study the coreference resolution problem, such as MUC \cite{muc-7}, ACE \cite{ace}, and OntoNotes \cite{conll12}. The most popular one is OntoNotes, and recent work on coreference resolution \cite{clark:16:rl, clark:16:entity, lee:17, peters:18} evaluated their models on it. Other datasets were rarely studied after OntoNotes was published.

Previous work \cite{Moosavi:17} suggests that the overlap between training and test sets makes significant impact on the performance of current coreference resolvers. In OntoNotes, which has relatively low training-test overlap, this impact is mixed together with the core challenges of coreference resolution. For example, consider the failure of referencing ``them'' to ``the wounded'' in ``..., the wounded were carried off so fast and it was difficult to count them''. It is hard to tell whether the algorithm can succeed if the currently low-frequency phrase ``the wounded'' has not been seen enough times in the training set.
From a machine learning perspective, high overlap is needed to ensure that the training and test datasets have similar statistics.

Another limitation of OntoNotes is that it only has annotations for non-singleton mentions, while singleton mentions are not annotated. Most of the algorithms for coreference resolution have two steps: mention detection and mention clustering \cite{wiseman2016global, clark:16:rl, clark:16:entity}. The lack of singleton mention annotations makes training and evaluation of mention detectors more difficult.

To address both limitations of OntoNotes, we build a new dataset, PreCo. To alleviate the negative impact of low training-test overlap, we restrict the data domain and collect a sufficient amount of data to achieve a relatively high training-test overlap.
Restricting the data domain is a common way to enable better studies of unsolved NLP tasks, such as language modeling \cite{hill2015goldilocks} and visual question answering \cite{johnson2017clevr}.

We select our data from English reading comprehension tests for middle and high school Chinese students, which has several advantages.
On one hand, the vocabulary size is appropriate. The English vocabulary of a typical Chinese high school student contains about 3000 commonly used words. This is similar to the vocabulary of a preschool English-speaking child \cite{vocabulary}. Most words from the English tests are in this limited vocabulary.
On the other hand, it is practical to collect enough data of this type from the Internet. With 12.4M words, PreCo is about 10 times larger than OntoNotes. Large scale datasets, e.g. ImageNet \cite{deng2009imagenet}, SQuAD \cite{rajpurkar2016squad}, have played an important role for driving computer vision and NLP forward.

We use the rate of out-of-vocabulary (OOV) words between training and test sets to measure their overlap. PreCo shows much higher training-test overlap than OntoNotes by having an OOV rate of 0.8\%, which is about 1/3 of OntoNotes's 2.1\%.
At the same time, PreCo presents a good challenge for coreference resolution research since its documents are in the open domain and have various writing styles.
We test a state-of-the-art system \cite{peters:18} on PreCo and get an F1 score of 81.5. However, a modest human performance (87.9, which will be described in \ref{sec:baseline} ) is much higher, verifying there remain challenges.

To help training and evaluation of mention detection, we annotate singleton mentions in PreCo. Besides singleton mentions, we follow most other annotation rules of OntoNotes to label the new dataset. We show that in a state-of-the-art coreference resolution system \cite{peters:18}, we can improve the model performance from 77.3 to 81.6 F1 on a training set of 2.5K PreCo documents by using an oracle mention detector, and the remaining gap of 18.4 F1 to the perfect 100 F1 can only be reduced by improving mention clustering. This indicates that future work should concern more about mention clustering than mention detection.

The advantages of our proposed dataset over existing ones in coreference resolution can be summarized as follows:
\begin{itemize}
\item Its OOV rate is about 1/3 of OntoNotes.
\item It has about 10 times larger corpus size than OntoNotes.
\item It has annotated singleton mentions.
\end{itemize}

\section{Related Work} \label{sec:related}

\textbf{Existing Datasets.}  The first two resources for coreference resolution study were MUC-6 and MUC-7 \cite{muc-7}. The MUC datasets are  too small for training and testing, containing a total of 127 documents with 65K words. The next standard dataset was ACE \cite{ace} which has a much larger corpus of 1M words. But its annotations are restricted to a small subset of entities and are less consistent. OntoNotes \cite{conll12} was presented to overcome those limitations. Machine learning based approaches, especially deep learning based, benefitted from this well annotated and large-scale (1.3M words) dataset. Continuous research on OntoNotes over the past 6 years improved performance by 10 F1 score \cite{durrett2013easy, peters:18}. Datasets after OntoNotes, such as WikiCoref \cite{ghaddar:16}, are seldom studied. Therefore, we mainly compare PreCo with OntoNotes in this paper. With a much larger scale, PreCo builds on the advantages of OntoNotes. Some of these existing datasets also have corpus in other languages, but we just focus on coreference resolution in English.

\textbf{Out-of-domain Evaluation.}  \cite{Moosavi:17} show that if coreference resolvers mainly rely on lexical representation, as it is the case in state-of-the-art ones, they are weak at generalizing to unseen domains. Even in the seen domains, the low degree of overlap for non-pronominal mentions between the training and test sets cause serious deterioration of coreference resolution performance. As a conclusion, \cite{Moosavi:17} suggested that out-of-domain evaluation is a must in the literature. But we think the problem can be relieved by expanding the training data for the target domains to increase overlap, so that the field can pay more attention to the other challenges of coreference resolution.

\textbf{Data Simplification.} Many simplified datasets were built to enable better study on unsolved tasks. Such simplifications can guide researchers to the core problems and make data collection easier. For example, \cite{hill2015goldilocks} introduced the Children's Book Test to distinguish the task of predicting syntactic function words from that of predicting low-frequency words for language model. The dataset helped them to develop a generalizable model with explicit memory representations. The reading comprehension dataset SQuAD \cite{rajpurkar2016squad} imposes the constraint that every answer is always a segment of the input text. This constraint benefits both labeling and evaluation of the dataset, which has significant influences in terms of benchmarks. Similarly, the reinforcement learning literature develops algorithms by studying games instead of the real world environment \cite{mnih2013playing}. We hope that, with high training-test overlap, PreCo can serve as a valuable resource for research on coreference resolution.

\section{Dataset Creation}

We discuss the data collection and annotation in this section. The overview of the process is shown in Figure ~\ref{fig:dataset_creation}.

\subsection{Corpus Collection}

We crawl English tests from several web sites. The web pages often contain the full English tests in a lot of formats.
We build an annotation website and hire annotators to manually extract the relevant contents.
We have a total of 80 part-time Chinese annotators, most of whom are university students. They are required to have a minimum score in standard English tests. During annotation training, the annotators read the annotation rules, and take several practice tasks, in which they annotate sample articles, and their results are compared with ground truth side by side for them to study. Before formal annotation, the annotators will need to pass an assessment.

Some data cleaning is done during annotation, such as unifying paragraph separators, etc. The questions with answers in these tests are also extracted for future research. Finally, we use NLTK's sentence and word tokenizer \cite{nltk} to tokenize the crawled text.

\begin{figure}[htbp]
\centering
\includegraphics[width=6.2cm]{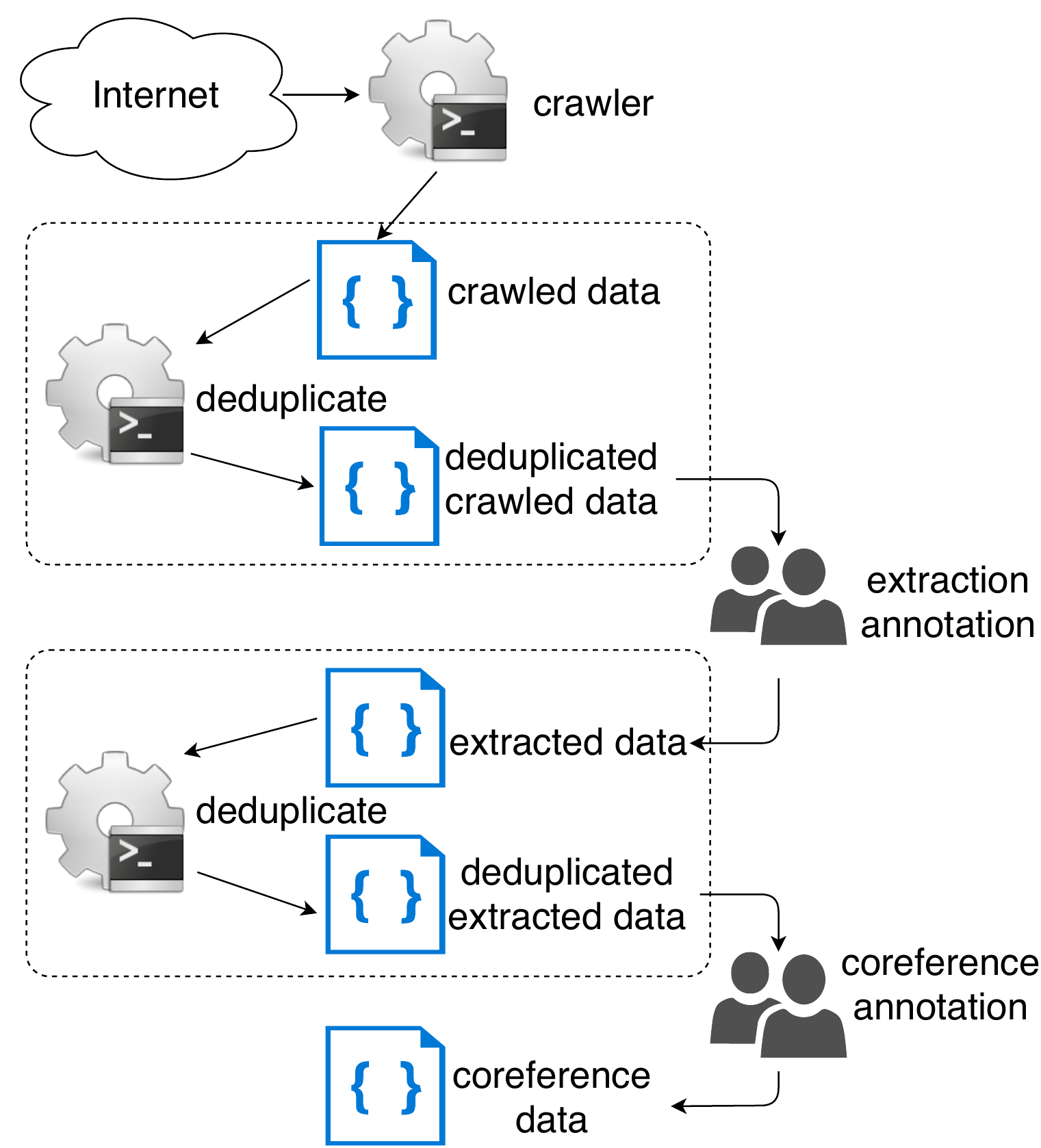}
\caption{Overview of dataset creation.}
\label{fig:dataset_creation}
\end{figure}

In addition to having annotators manually clean the data, we also use heuristic rules to further clean the data. For example, in some cases the whitespaces between two words are missing. We use a spell checker to identify and correct most of these cases. We also use heuristic rules to fix some sentence partition boundaries, e.g., to make sure opening quotes are placed at the beginning of a sentence, instead of being wrongly placed at the end of a previous sentence (closing quotes are handled similarly).

In addition to the crawled data, we include the documents from the RACE dataset \cite{lai:17race}. RACE is a reading comprehension dataset from English tests for middle and high school Chinese students, which has similar types of data sources as PreCo. About 2/3 of PreCo documents are from the RACE dataset.

Since documents are from several data sources, we want to remove duplicated documents, and documents that are not exactly the same but have a high rate of repetitions.
The similarity of two documents $D_1$ and $D_2$ is estimated using the bag-of-words model. Assume $S_1$ and $S_2$ are bag-of-words multisets to represent the two documents. The similarity between $D_1$ and $D_2$ is defined as $\max({\frac{|S_1 \cap S_2|}{|S_1|}, \frac{|S_1 \cap S_2|}{|S_2|}})$. If the similarity between two documents are larger than 0.9, we remove the shorter one. This process is referred as \emph{deduplicate} in Figure \ref{fig:dataset_creation}.

\subsection{Data Partition}

The dataset has a total of 37.6K documents. We use 500 documents for the development set, 500 documents for the test set, and the rest 36.6K documents for the training set. The development and test documents were randomly selected from RACE's development and test sets.

\subsection{Coreference Annotation and Refinement} \label{sec:anno_and_refinement}

We manually annotate coreferences on these documents. The annotation rules are slightly different from OntoNotes \cite{conll12}. We modify some of the rules to make the definition of coreference more consistent and easier to be understood by the annotators. The major differences are listed in Table \ref{tab:anno_rules}. Figure \ref{fig:example_text} shows an example document in PreCo with annotations.

\begin{table*}[htbp]
\centering

{
\small
\begin{tabular}{L{1.5cm} L{4cm} L{4.5cm} L{4.5cm}}
\toprule
{\bf Type} & {\bf Example} & {\bf OntoNotes} & {\bf PreCo} \\

\midrule
verbs &
Sales [grew] 10\%. [The growth] is exciting. &
Verbs can be coreferred. &
Usually, verbs cannot be coreferred. Certain gerunds can. \\

\midrule
generic mentions &
[Parents] are usually busy. [Parents] should get involved. &
Generic mentions can only be coreferred by pronouns. &
Generic mentions can be coreferred directly. \\

\midrule
non-proper modifiers &
[Wheat] is important. [Wheat] fields are everywhere. &
Non-proper modifiers cannot be coreferred. &
Non-proper modifiers can be coreferred as generic mentions. \\

\midrule
copular structures &
[John] is [a good teacher]. &
The referent and the attribute cannot be coreferred. &
The referent and the attribute can be coreferred. \\

\midrule
appositives &
[[John]$_a$, [a linguist I know]$_b$]$_c$, ... &
Sub-spans are not coreferred with the whole-span. $a$ and $b$ are not coreferent with $c$. &
Sub-spans are coreferred with the whole-span. $a$ and $b$ are coreferred with $c$. \\

\midrule
misc. &
The [U.S.] policy ... [Secretary of State] [Colin Powell] ... &
Nationality acronyms and job titles in appositives cannot be coreferred. &
Nationality acronyms and all job titles can be coreferred.\\

\bottomrule
\end{tabular}
}
\caption{Major differences of annotation rules between PreCo and OntoNotes. The annotation rules of OntoNotes are described in \cite{CoNLL12rules}}
\label{tab:anno_rules}
\end{table*}

\begin{figure}[h]
\centering
\includegraphics[width=4.5cm]{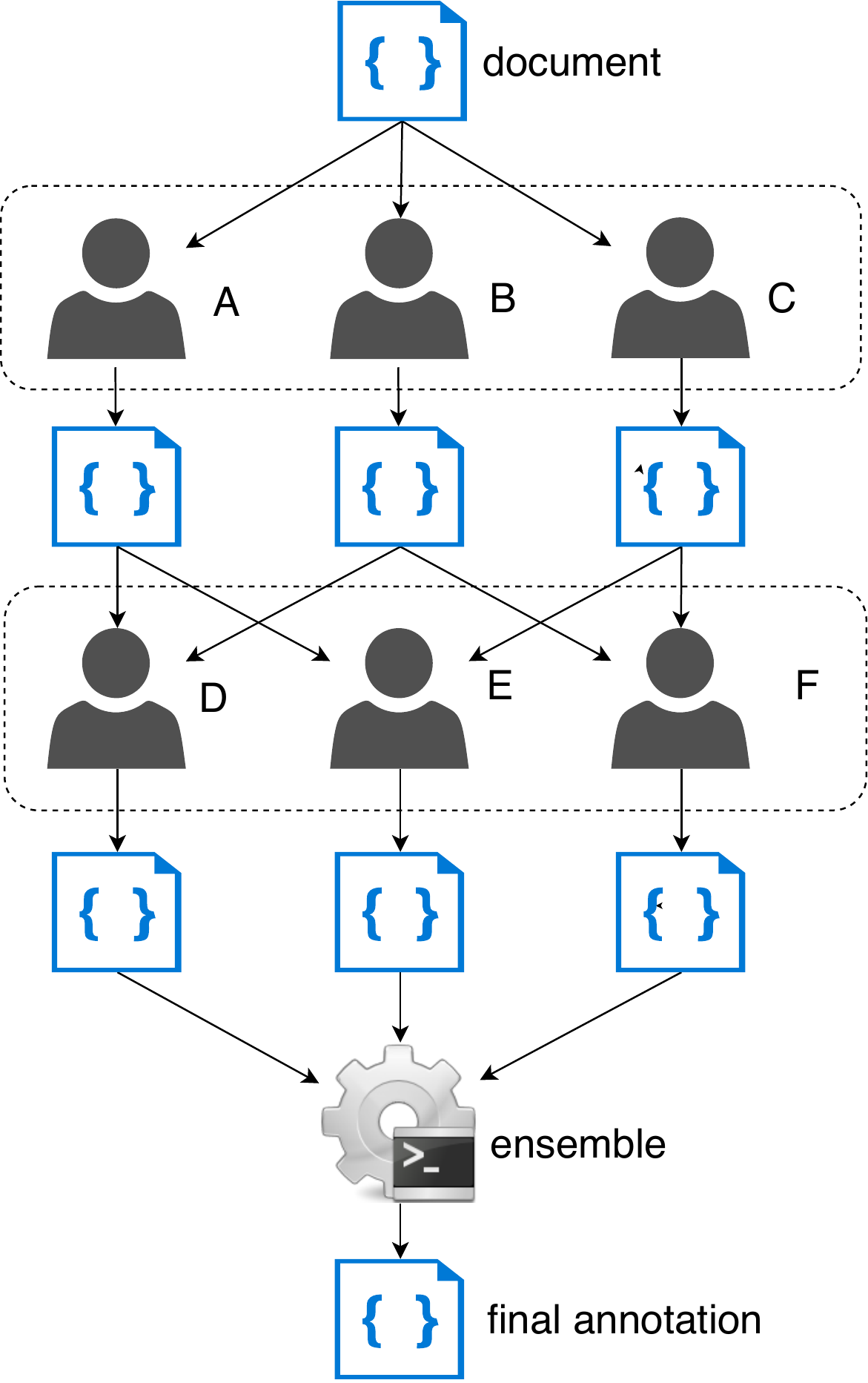}
\justify
\caption{Process of annotation refinement. A document is firstly annotated by 3 annotators A, B, and C, independently. Then another annotator D  merges annotations from A and B. Similarly, annotator E merges annotations from A and C, and annotator F merges annotations from B and C. Finally, annotations from D, E and F are merged using an ensemble algorithm.}
\label{fig:refine_anno}
\end{figure}

Good quality control of annotation is essential, since the rules are complicated and coreference resolution depends on meticulous reading of the whole document over and over. We found that annotators get low recall and insufficient precision mainly because of negligence, as opposed to the lack of annotation rules or other ambiguities. For example, two co-referred mentions could be far apart and require careful searches, and an annotator may miss it.
Therefore we further refine annotations as shown in Figure \ref{fig:refine_anno}. Annotators can think about the complicated inconsistent cases when merging annotations, and the voting process will fix some errors while preserving the mentions and coreferences that are found only once by individual annotators.

The quality of different annotation processes is shown in Table \ref{tab:anno_quality}. OntoNotes took 2 individual annotations for each document and got an adjudicated version based on them. Taking the adjudicated version as ground truth, the average MUC score \cite{vilain1995MUC} \footnote{MUC score is one of the metrics to evaluate the quality of coreference resolution.} of individual annotations is 89.6, and the inter-annotator MUC score is 83.0. The corresponding numbers for PreCo are 85.3 and 77.5. The actual gap of individual annotation quality between OntoNotes and PreCo is not as large as it looks like. Note that, OntoNotes's two individual coreference annotations of each document are based on the same syntactic annotations of the document, so they could be more consistent than PreCo's which are annotated on raw text. Therefore, if we want to fairly compare PreCo with OntoNotes, we should take into account OntoNotes's inter-annotator consistency of syntactic parsing annotations. As it has a rough upper bound of 98.5 F1 score according to the reannotation of English Treebank on OntoNotes by the principal annotator a year after the original annotation \cite{weischedel2011ontonotes}, we could infer that the individual annotation quality of PreCo is quite close to OntoNotes.

Labeling the whole dataset is costly because each annotation from scratch or comparison takes an average of about 10 minutes. Prompts from an algorithm do not help since they do not speed up the annotation much but instead introduce biases. We observed some biases when using an algorithm to help annotation. We have two models, $M_1$ and $M_2$, and we have a test set $T$ which is annotated manually, and a test set $T'$ which uses prompts from model $M_1$ to help annotation. While $M_1$ and $M_2$ have similar performance on $T$, $M_1$'s performance is much higher than $M_2$'s on $T'$, which shows the biases.

\begin{table}[h]
\centering
\begin{tabular}{l c c c}
\toprule
{\bf Process} & \makecell{\bf Avg. Prec} & \makecell{\bf Avg. Rec} & \makecell{\bf Avg. F1} \\
\midrule
Once & 87.3 & 71.7 & 78.7 \\
ABC-voting & 93.5 & 76.1 & 83.9 \\
AB-merge & 87.5 & 88.3 & 87.9\\
DEF-voting & 100.0 & 100.0 & 100.0 \\
\bottomrule
\end{tabular}
\justify
\caption{Annotation quality. DEF-voting is taken as the ground truth to evaluate other annotation processes. The annotation ``AB-merge'' is merged by annotator G, who is different from D, E and F.}
\label{tab:anno_quality}
\end{table}

Because of limited annotation resources, we have only finished the refinements on the development and test sets with the process shown in Figure \ref{fig:refine_anno}. We refine the training set annotations as follows: for each document, two annotators annotate it separately, and a third annotator compares and merges the two annotations. We use a training set of 2.5K documents to quantify the impact of this annotation refinement to model performance. Table \ref{tab:anno_quality_impact} shows the model performances of the training set that is annotated once, and the training set of the merged annotation. The performance difference is quite significant. Furthermore, the difference is consistent with Table \ref{tab:anno_quality}: the ``AB-merge'' model has a similar precision as the ``Once'' model, but it has a much higher recall. It indicates that a further refinement of the training set such as DEF-voting could be essential. A more interesting question is: how to make the definition of coreference more consistent and executable? We leave it as future work.

\begin{table}[h]
\centering
\begin{tabular}{l c c c}
\toprule
{\bf Annotation} & \makecell{\bf Avg. Prec} & \makecell{\bf Avg. Rec} & \makecell{\bf Avg. F1} \\
\midrule
Once & 79.3 & 69.1 & 73.9 \\
AB-merge & 78.1 & 76.5 & 77.3\\
\bottomrule
\end{tabular}
\justify
\caption{The annotation quality's impact on model performance. Each row shows the development set performance of the EE2E-Coref model (training details in Section \ref{sec:baseline}) trained by data of different annotation quality. Each training set contains 2.5K documents. In the training set ``Once'', each document is annotated by one annotator. In the training set ``AB-merge'', each document is annotated by two annotators independently, and the annotations are compared and merged by a third annotator.}
\label{tab:anno_quality_impact}
\end{table}

\subsection{Dataset Properties} \label{sec:dataset_properties}

Table \ref{tab:dataset_properties} shows some properties of OntoNotes and PreCo. As intended, PreCo has a lower OOV rate than OntoNotes. For a training set with vocabulary $\mathcal{V}$ and a test set with $n$ tokens $[t_1, t_2, ..., t_n]$, ignoring the tokens with non-alphabetic characters, the OOV rate is defined by:
$$\frac{\sum_{i}{o(t_i)}}{n}, \text{where } o(t_i)=
\begin{cases}
0 & \text{if } t_i \in \mathcal{V} \\
1 & \text{if } t_i \notin \mathcal{V}
\end{cases}$$
The OOV rate can be extended to the rate of low-frequency words which also indicates the training-test overlap, by simply replacing $\mathcal{V}$ in the definition above with the non-low-frequency vocabulary of the training set. We find that the OOV rate is consistent to the rates of low-frequency words in different levels. So we use the OOV rate for convenience.

\begin{figure}[htbp]
\centering
\includegraphics[width=7cm]{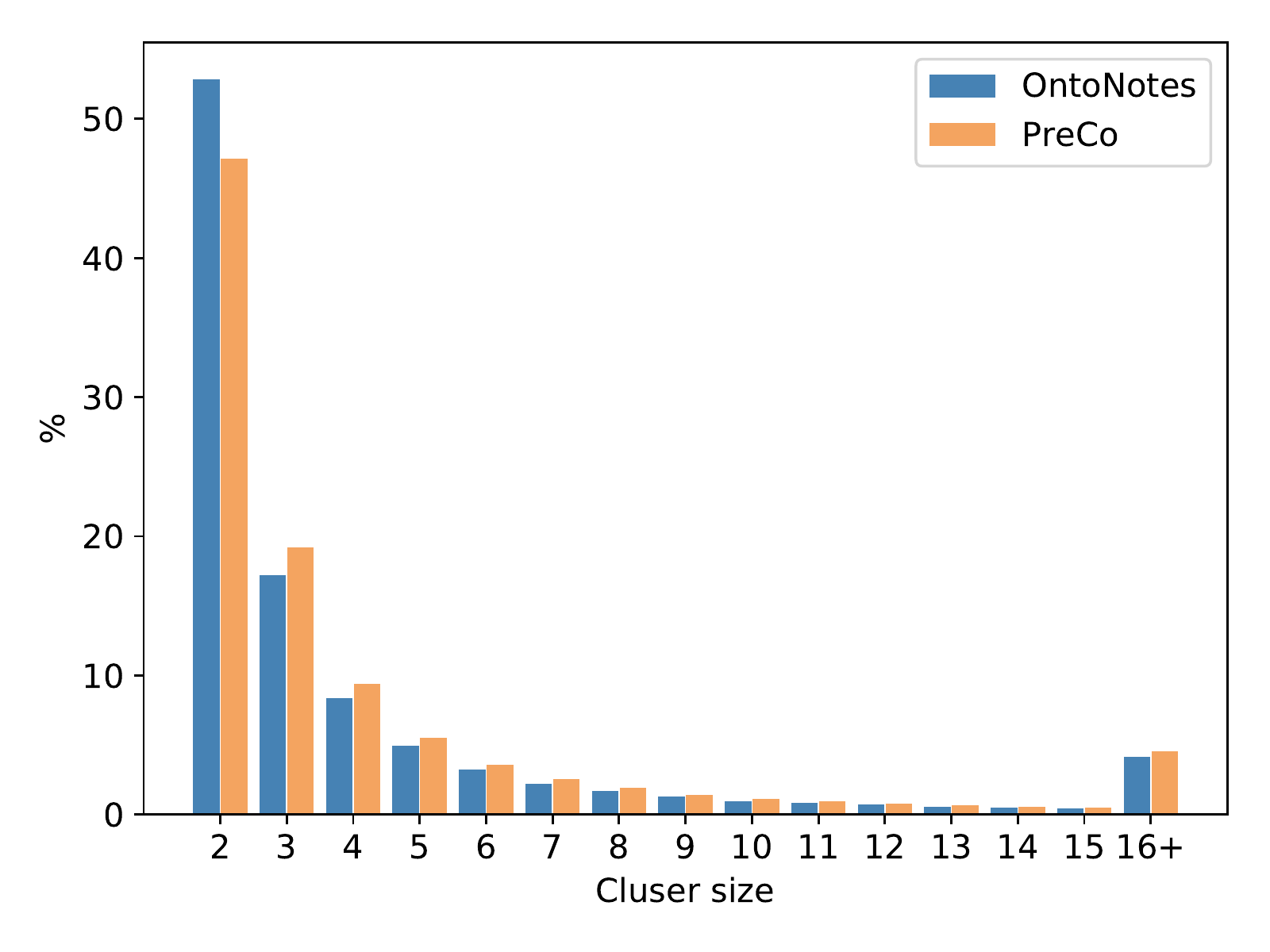}
\caption{Distribution of cluster sizes within non-singleton clusters. We ignore singleton clusters in this figure so that it is easier to compare between OntoNotes and PreCo.}
\label{fig:cluster_size_distribution}
\end{figure}

In PreCo, about 50.8\% of the mentions are singleton mentions. Figure \ref{fig:cluster_size_distribution} shows the distribution of cluster sizes within non-singleton clusters. The distribution is similar between OntoNotes and PreCo.

\begin{table}[h]
\centering

\begin{tabular}{l c c}
\toprule
{\bf Property} & {\bf OntoNotes} & {\bf PreCo} \\
\midrule
Training documents     & 2.8K  & 36.6K \\
Training tokens        & 1.3M  & 12.1M \\
Dev-test documents     & 0.7K  & 1K \\
Dev-test tokens        & 0.3M  & 0.3M \\
Tokens per document    & 467   & 330 \\
OOV rate               & 2.1\% & 0.8\% \\

\midrule
\multicolumn{3}{c}{Non-singleton mentions} \\
\midrule
Mention length          & 2.29  & 2.02 \\
Mention density         & 0.12  & 0.16 \\
Cluster size            & 4.40  & 4.49 \\
Cluster density         & 0.027 & 0.035 \\

\midrule
\multicolumn{3}{c}{Singleton mentions} \\
\midrule
Mention length          & N/A   & 3.32 \\
Mention density         & N/A   & 0.16 \\
Singleton mention rate  & N/A   & 50.8\% \\

\bottomrule
\end{tabular}

\justify
\caption{Properties of OntoNotes and PreCo. The mention (cluster) density is defined by: number of mentions (clusters) / number of tokens.}
\label{tab:dataset_properties}
\end{table}

\section{Analysis}

To verify our assumption that PreCo embodies the core challenges of coreference, we evaluate a strong baseline coreference resolver on it. Specifically, we (i) estimate the room for improvement of the baseline system to show that the dataset is challenging, (ii) study the impact of training-test overlap to model performance and error analysis to show the advantages of PreCo, and (iii) quantitatively evaluate the mention detector to understand the bottlenecks of the coreference resolution system.

\subsection{Baseline Performance}\label{sec:baseline}

We use the end-to-end neural coreference resolver, E2E-Coref \cite{lee:17}, enhanced by the deep contextualized word representations \cite{peters:18} as the baseline system, and we refer to this system as EE2E-Coref. This is the state-of-the-art model on OntoNotes, achieving a test average F1 score of 70.4, which is the main evaluation metric for coreference resolution. The metric is computed by averaging the F1 of MUC, B$^3$, and CEAF$_{\phi 4}$, which are three metrics of coreference resolution that have different focuses.

Our implementation EE2E-Coref\footnote{It gets an F1 score of 70.0$\pm$0.3 on OntoNotes, slightly lower than the F1 score reported in the original paper.} gets 81.5 Avg. F1 score on PreCo. We follow the setting of most hyperparameters on OntoNotes and do grid-search for the decay parameter of the learning rate and the size of the hidden layers on the development set, since these two hyperparameters are relatively sensitive to the scale of the training data. The F1 score increment from OntoNotes to PreCo is probably due to the higher overlap between the training and test sets in PreCo.

\begin{table}[h]
\centering

{\small
\begin{tabular}{L{7.2cm}}
\toprule
{[}\textless His father\textgreater \ and he{]} get off the car. \\
{[}They{]} find the old man lying near the taxi. \\
The banana skin is near him. \\
The old man looks at {[}them{]} and says, ``Teach {[}\textbf{your}{]} child to throw the banana skin to the right place!'' \\
\midrule
He gave his last few coins to {[}a beggar{]}, but then he saw \textless another one\textgreater, and forgot that he did not have any money. \\
He asked \textless the man\textgreater \ if \textless he\textgreater \ would like to have lunch with him, and {[}\textbf{the beggar}{]} accepted, so they went into a small restaurant and had a good meal. \\
\midrule
{[}Holmes{]} and \textless Dr. Watson\textgreater \ went on a camping trip. \\
After a good meal and a bottle of wine, they lay down in a tent for the night and went to sleep. \\
Some hours later, {[}Holmes{]} woke up and pushed \textbf{his friend}. \\
\bottomrule
\end{tabular}
}

\justify
\caption{Error cases of EE2E-Coref on PreCo. Each bold mention is incorrectly referred to the entity in {[}{]}s. The mentions of its gold entity are in \textless \textgreater s.}
\label{tab:error_case}
\end{table}

We demonstrate three typical error cases made by EE2E-Coref on PreCo in Table \ref{tab:error_case}. Coreference resolution in these cases requires good understanding of multiple sentences, which is an open problem in NLP. A capable entity representation for ``them'', ``another one'' or ``Dr. Watson'' may help to resolve these error cases. We also compare the performance of EE2E-Coref with human performance to estimate the room for improvement on PreCo. As described in Section \ref{sec:dataset_properties}, human annotators get low recall mostly due to negligence. So we use the AB-merge annotation to estimate human's ability on coreference resolution. The gap of performance between model and human is 6.4 F1 score, from 81.5 to 87.9. The actual gap is larger, since AB-merge still has some missed coreference annotations due to negligence. This shows that the dataset is challenging and encourages future research. The error cases show the challenges as well.

Note that PreCo is not a general purpose dataset. Our motivation of designing PreCo is to make it easier to improve coreference resolution algorithms, e.g., to make error analysis easier. It is not a goal of PreCo to generalize well on corpus from other domains. Furthermore, we find that there are a certain amount of annotation errors in the development and test sets. We suggest that researchers working on PreCo should be careful about these errors, especially after a model gets F1 score beyond 90.0.

\subsection{Impact of Training-test Overlap} \label{sec:oov_impact}

Training-test overlap makes significant impact on error analysis. Consider an error case of coreference resolution, if there are low-frequency words in the related mentions, then it will be hard to tell whether the algorithm can succeed if the words has not been seen enough times in the training set. We call an error case LFW if there are low-frequency words\footnote{In our experiments, a word is defined as low-frequency if it appears in the training set less than 10 times.} in its related mentions\footnote{There are 3 kinds of error cases of coreference resolution: false-new, false-link and wrong-link. In our experiments, the related mentions include: the current mention in all 3 kinds of cases, the nearest gold antecedent in false-new and wrong-link and the false referred antecedent in false-link and wrong-link.}. Therefore, the lower LFW rate a training set contains, the more precisely it may expose the drawbacks of the algorithm.

\begin{figure*}[htbp]
\centering
\subfigure[]{
\label{fig:oov_impact_ds_f1}
\begin{minipage}{6.9cm}
\centering
\includegraphics[width=6.9cm]{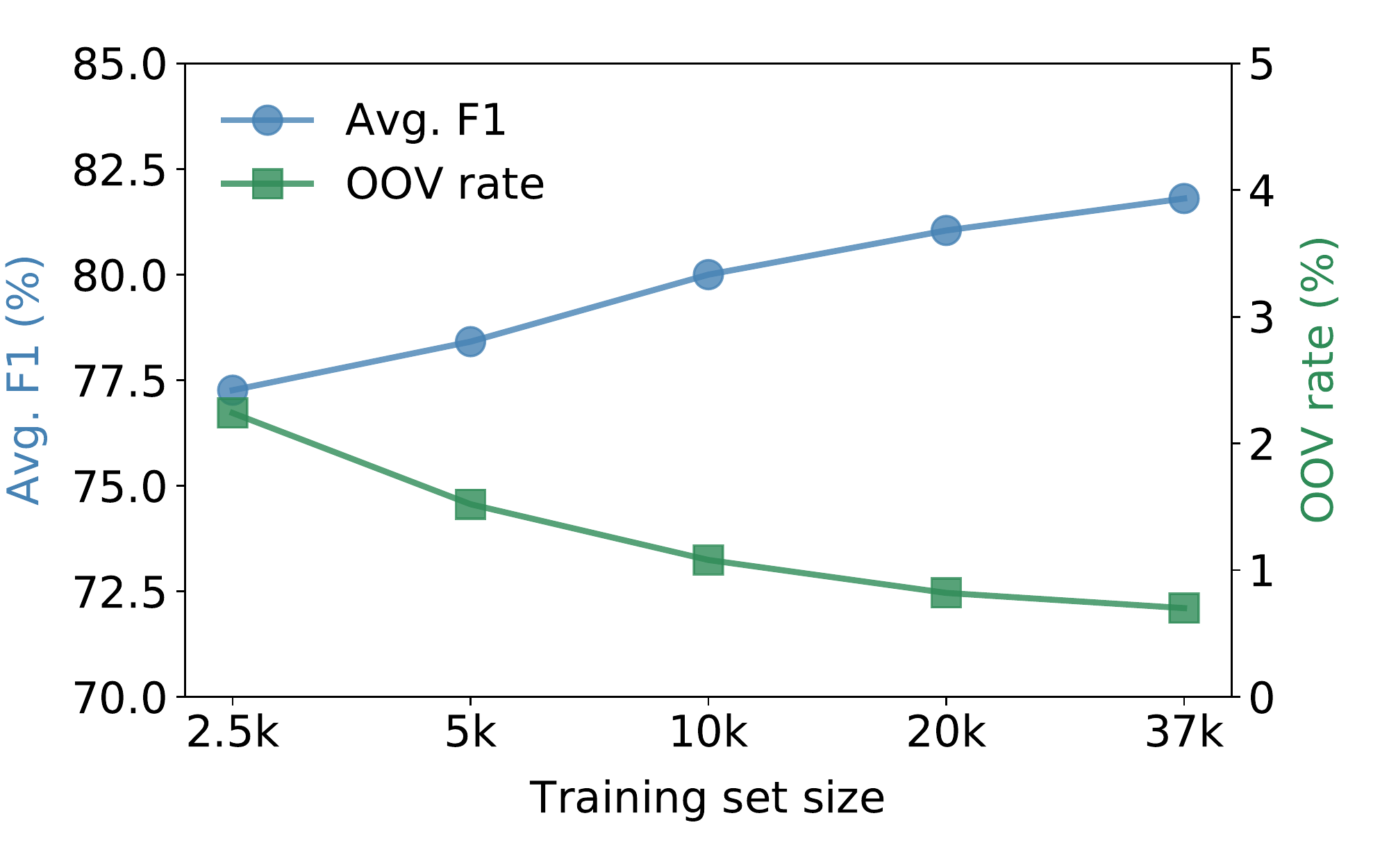}
\end{minipage}}
\subfigure[]{
\label{fig:oov_impact_ds_lfw}
\begin{minipage}{6.9cm}
\centering
\includegraphics[width=6.9cm]{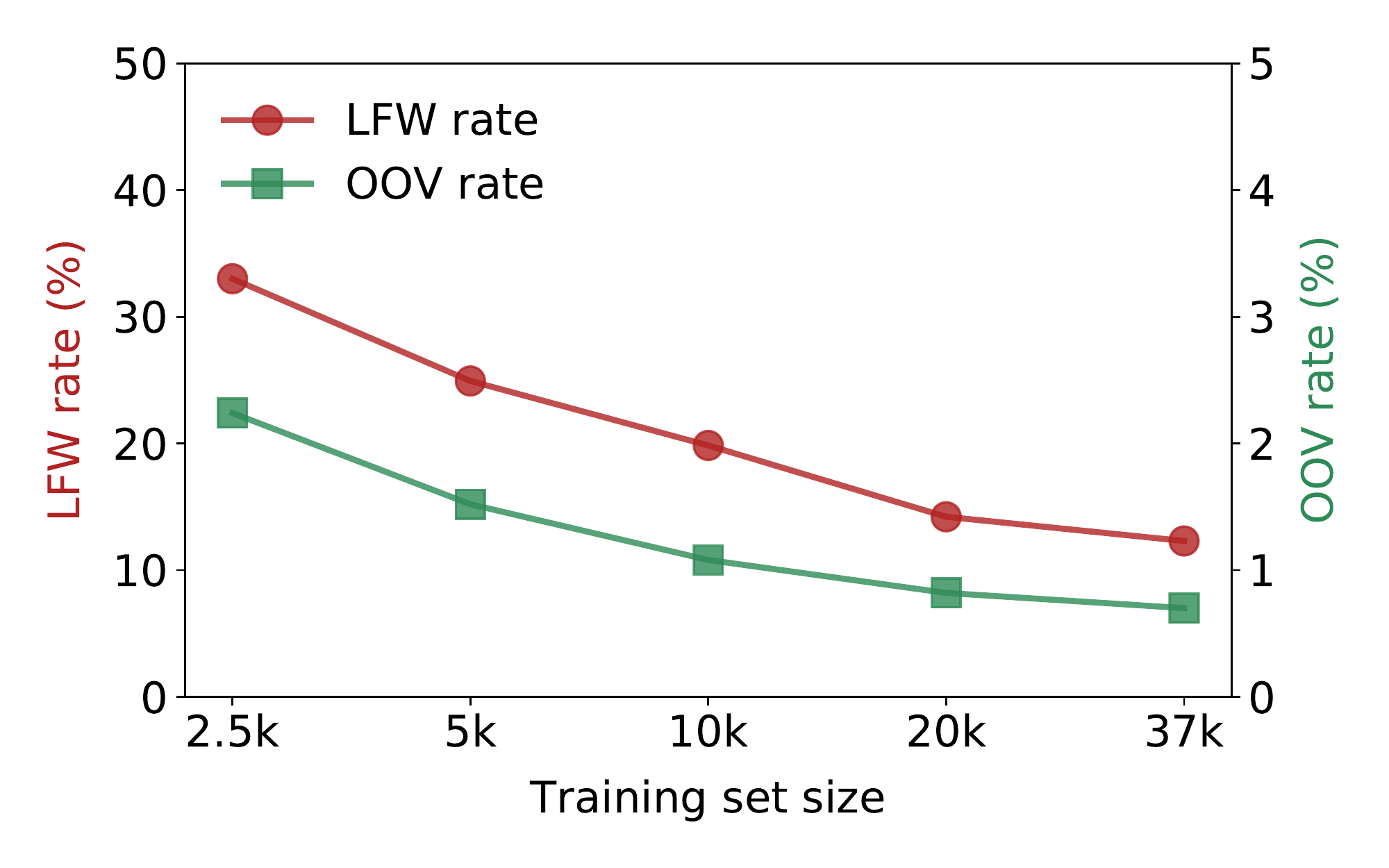}
\end{minipage}}
\subfigure[]{
\label{fig:oov_impact_ss_f1}
\begin{minipage}{6.9cm}
\centering
\includegraphics[width=6.9cm]{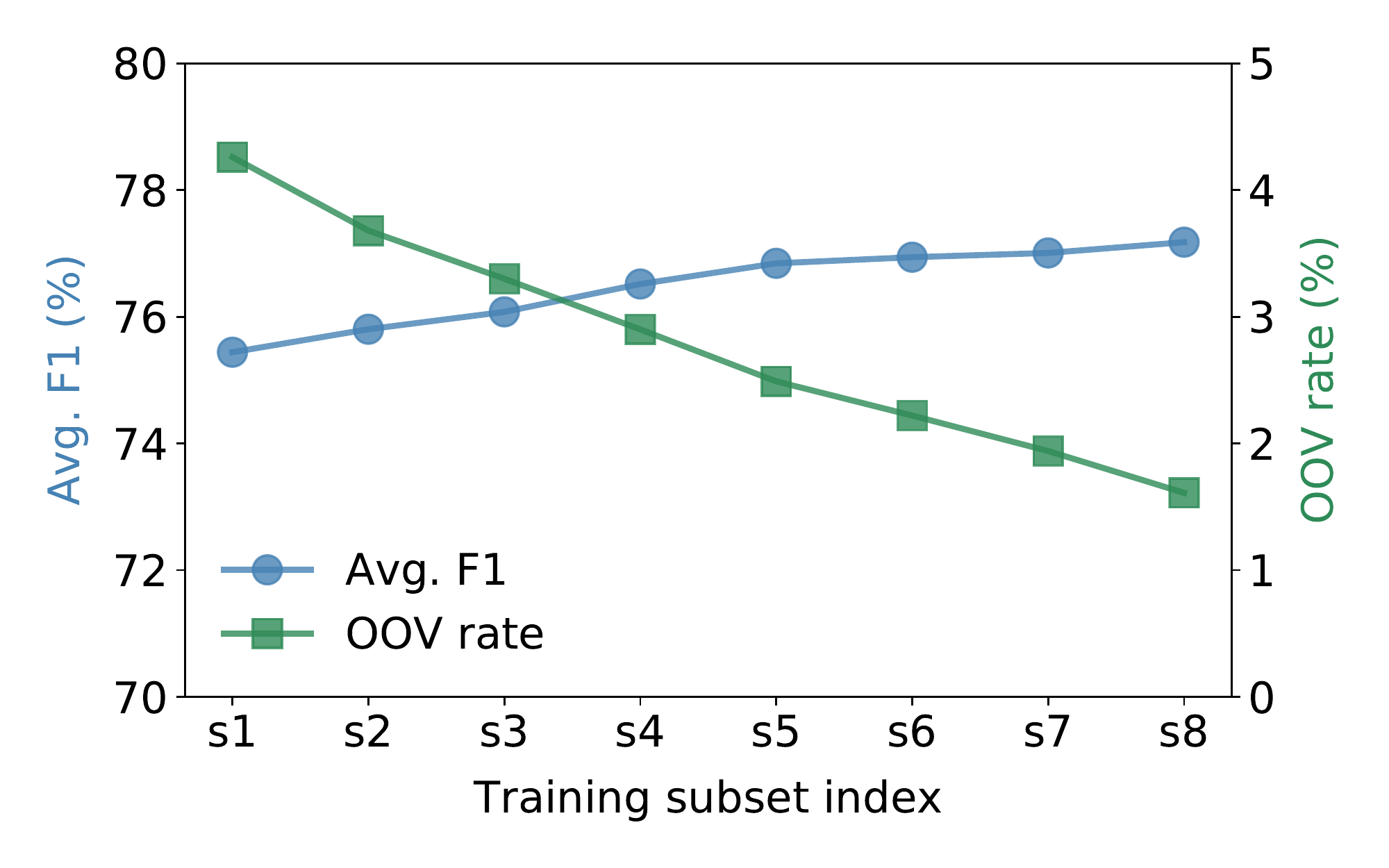}
\end{minipage}}
\subfigure[]{
\label{fig:oov_impact_ss_lfw}
\begin{minipage}{6.9cm}
\centering
\includegraphics[width=6.9cm]{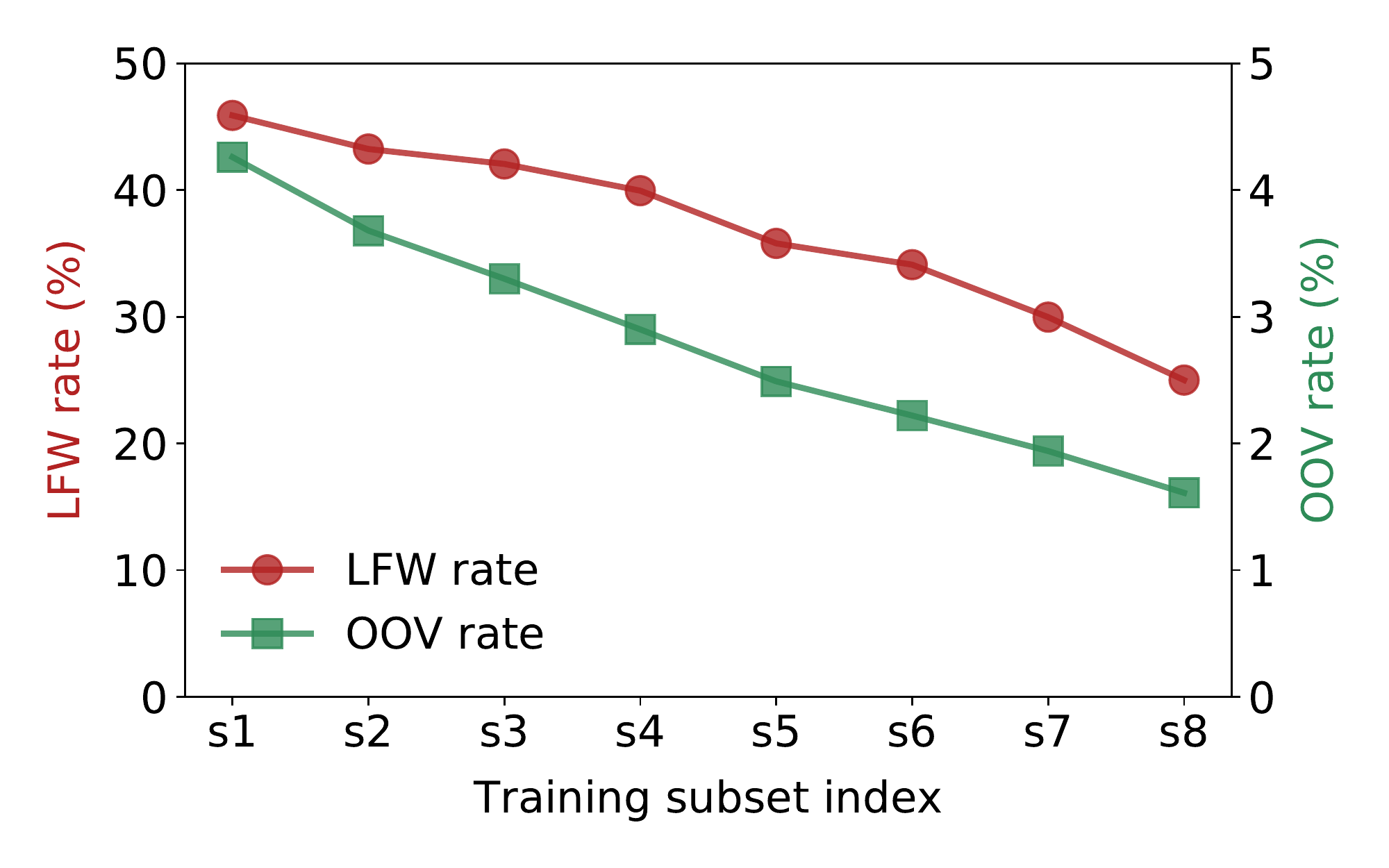}
\end{minipage}}
\\
\caption{Impact of training-dev overlap. (a) and (b) show the impact of training set sizes. (c) and (d) show the impact of the training-dev OOV rate, when the training sets have the same size of 2.5K documents. The 8 subsets, s1-s8, consist of documents ranked by their overlaps with the development set vocabulary.}
\end{figure*}

To study the impact of training-test overlap, actually, the training-dev overlap, we pick different subsets from the training data and evaluate the models trained on them. At first, we control overlap by picking different sizes of the training data randomly. Figure \ref{fig:oov_impact_ds_f1} shows that, as the training data size grows, the OOV rate, which is the overlap indicator, decreases and the F1 score of EE2E-Coref increases significantly. Figure \ref{fig:oov_impact_ds_lfw} shows that when training set size increases, the OOV rate and the LFW rate drop together. Then, to remove the impact of data size, we pick training sets which have a fixed size but different overlaps with the development set vocabulary. The OOV rates and F1 scores of these subsets are shown in Figure \ref{fig:oov_impact_ss_f1}. This experiment verifies the positive correlation between training-dev overlap and coreference resolution performance suggested by \cite{Moosavi:17}. Figure \ref{fig:oov_impact_ss_lfw} shows that for training sets with the same size, the OOV rate and the LFW rate also drop together.

We observe that the training set of 2.5K documents in Figure \ref{fig:oov_impact_ds_f1} has a higher model performance than all the training sets in Figure \ref{fig:oov_impact_ss_f1}. This is not expected. One hypothesis is that the lower performance in Figure \ref{fig:oov_impact_ss_f1} is due to the smaller diversity of these training sets, which are selected to have certain training-dev OOV rates.

The training-dev LFW rate of OntoNotes is 34.8\%. As a comparison, the number for PreCo is 12.3\%. A subset of PreCo with a similar token number to OntoNotes has a LFW rate of 33.0\%. This indicates that research of coreference algorithms on PreCo will be much more efficient than on OntoNotes. Even if we can ignore the LFW error cases, there are others related to low-frequency word senses, phrases and sentence structures, which are hard to filter out. They will also obscure the error analysis. It is reasonable to believe that training-dev overlap impacts the rate of these error cases in a similar way to impact LFW rate.

\subsection{Mention Detection} \label{sec:mention_detection}

Since most coreference systems consist of a mention detection module and a mention clustering module, an important question is: with a perfect mention detection module, what is the model performance on coreference resolution? The answer would help us understand the bottlenecks of the entire system, by quantifying the impact of the mention detection module on the final F1 score. \cite{lee:17} gave an answer by taking ground truth non-singleton mentions as the input of the coreference resolver for both training and evaluation, assuming that the perfect mention detector can also make perfect anaphoricity decisions, e.g., to decide whether a mention should be linked to an antecedent. But this assumption can be violated since mention detectors usually take local information but anaphoricity decisions usually need more context, nearly as much as entity identification. The anaphoricity decisions should be made in the mention clustering module.

\begin{table}[h]
\centering
\begin{tabular}{l c c}
\toprule
{\bf Mention} & {\bf OntoNotes} & {\bf PreCo} \\
\midrule
detected & 66.7 & 77.3 \\
*all & N/A & 81.6 \\
*non-singleton & 85.2 & 89.2 \\
\bottomrule
\end{tabular}
\justify
\caption{Coreference resolution performances on development set under different mention detection qualities. A prefixed * denotes ground truth. The model trained on OntoNotes is E2E-Coref \cite{lee:17} while the one trained on PreCo is EE2E-Coref. The PreCo training set contains the same 2.5K documents as in Table \ref{tab:anno_quality_impact}. }
\label{tab:men_det_impact}
\end{table}

We argue that a better way to answer the question is to take all ground truth mentions (including singletons) for coreference. This operation is not feasible in OntoNotes since it does not have annotations for singleton mentions. We do this on PreCo and the results are shown in Table \ref{tab:men_det_impact}. There is an obvious difference between the F1 scores achieved with all gold mentions and non-singleton gold mentions. Therefore, the room for improvement by better mention detection is not as enormous as suggested in \cite{lee:17}. The major challenge remained in coreference resolution is mention clustering.

\section{Conclusion}
In this paper, we propose a large-scale coreference resolution dataset to overcome the limitations of existing ones. Our dataset, PreCo, features higher training-test overlap, about 10 times larger scale than previous datasets, and singleton mention annotations. By evaluating a state-of-the-art coreference resolver, we show that there is a wide gap between the model and human performance, which demonstrated challenges of the dataset. We verified the expectation that PreCo's higher training-test overlap helps research on coreference resolution. For the first time, we quantified the impact of mention detector to the entire system, thanks to our singleton mention annotations. We make the dataset public, and hope it will stimulate further research on coreference resolution.

\bibliography{PreCo}
\bibliographystyle{acl_natbib_nourl}

\end{document}